\documentclass{article} 
\usepackage{iclr2027_conference,times}

\usepackage{amsmath,amsfonts,bm}

\def\eqref#1{equation~\ref{#1}}

\def\1{\bm{1}}

\DeclareMathAlphabet{\mathsfit}{\encodingdefault}{\sfdefault}{m}{sl}
\SetMathAlphabet{\mathsfit}{bold}{\encodingdefault}{\sfdefault}{bx}{n}

\usepackage[table]{xcolor}
\definecolor{avgcol}{RGB}{234,242,255}
\usepackage{wrapfig}
\usepackage{hyperref}
\usepackage{url}
\usepackage{booktabs}
\usepackage{graphicx}
\usepackage{multirow}
\usepackage{float}
\usepackage{array}
\usepackage{longtable}
\usepackage{xspace}
\usepackage[textwidth=2.5cm,size=tiny]{todonotes}
\usepackage{enumitem}
\usepackage{verbatim}
\usepackage{microtype}

\newcommand{\method}{\textsc{SERA}\xspace}

\title{Self-Evaluating Recursive Agents}

\author{ Tianyi Lyu$^{1}$$^{*}$, Xiaozhe Li$^{1}$$^{*}$, Yang Li$^{2}$, Yongkang Chen$^{3}$, Kefei Tian$^{4}$, \bf Junbo Niu$^{5}$, \\ \bf Zican Hu$^{6}$, Hongbo Liu$^{1}$, Mingliang Xiong$^{1}$, Qingwen Liu$^{1}$$^{\dag}$\\
Tongji University$^{1}$, Independent$^{2}$, Shanghai Artificial Intelligence Laboratory$^{3}$, \\
Shanghai Jiaotong University$^{4}$, Peking University$^{5}$, Nanjing University$^{6}$\\
\url{https://github.com/OliverLeeXZ/SERA} \\
}

\iclrfinalcopy 
\begin{document}

\maketitle
\footnotetext[1]{$*$ represents equal contribution}
\footnotetext[2]{$^{\dag}$ represents Corresponding authors.}

\begin{abstract}
Recursive language-model agents decompose tasks and delegate subtasks to child instances of the same policy, forming a tree of work. Training them, however, is hard: the final outcome is verifiable, but the self-invented intermediate subtasks are numerous and carry no ground truth. Existing methods score each node with a verifier or judge, which is costly at scale and blind to decomposition quality. We argue that a recursive agent must learn three coupled capabilities within one set of weights: decomposing problems into subtasks, solving them, and evaluating the outcomes, each requiring its own training signal. SERA (Self-Evaluating Recursive Agents) turns evaluation into a learned capability of the policy itself. Before delegating, the parent writes a rubric of weighted success criteria for each child subtask; a ranking objective against verified outcomes then trains rubric generation so that the criteria track genuine subtask success. In addition, a complementary leaf-coverage signal provides direct credit for task decomposition. Our central finding is that \emph{training} the policy to generate aligned rubrics is what drives the gains: because the same weights both evaluate and execute, learning to judge subtasks sharpens the agent's ability to solve them. Notably, external supervision is also reduced: the judge is consulted only to train the rubric generator, while solving is trained against the agent's own rubric scores, which outperform direct use of the judge. Beyond training, the learned rubric doubles as an inference-time selector for tree search. On TextCraft-Synth and TextWorld-Sync, SERA improves over strong recursive-agent baselines by $5.38$ and $13.14$ points on average, and rubric-guided tree search at inference adds a further $2.43$ points on TextWorld-Sync.\looseness-1
\end{abstract}

\vspace{-5pt}
\section{Introduction}
\label{sec:intro}
\vspace{-3pt}


Language-model agents are deployed on tasks that cannot be solved in a pass: multi-hop research questions require coordinating evidence across retrieval steps; repository-scale code changes involve interdependent edits. A natural response is to decompose the problem and delegate each piece to a child agent \citep{prasad2024adapt, schroeder2025thread, chi2025era}, turning one query into a tree of work where each child handles a focused subproblem in a fresh context. When the children are instances of the \emph{same} policy, the system is \emph{recursive} \citep{zhang2025recursive, gandhi2026recursive}: a single set of weights drives every node. This is the design behind frontier coding and research agents \citep{claudecode, codex, tie2026autoresearch}, and it composes across depth: a policy trained at shallow recursion generalises to deeper trees at inference \citep{gandhi2026recursive}, making recursion a distinct axis of test-time scaling alongside longer chains of thought and wider sampling.

Training recursive agents is hard because supervision is scarce and unevenly distributed across the tree. The final outcome is typically verifiable (a research answer can be checked, code can be tested), but the intermediate subtasks the agent invents rarely admit a ground-truth check. The natural remedy is to score each node by its own success instead of broadcasting the root reward, using a verifier or an LLM judge (e.g., \citealp{gandhi2026recursive}). This localises credit, but the resulting signal has three limitations. It is \emph{coarse}: a binary per-node signal cannot separate a clean solution from a barely adequate one, or flag weak work inside a tree that still succeeds. It is \emph{expensive}: a reliable signal at every node needs a verifier or judge call throughout the tree, which scales poorly. And it is \emph{silent about decomposition}: rewarding whether subtasks were solved says nothing about whether they were the \emph{right} subtasks to create, though that choice often decides whether the tree succeeds.

The root issue is structural: a recursive agent is not doing one thing but three. The same shared weights must \emph{decompose} hard problems into tractable pieces, \emph{solve} the subtasks that result, and \emph{evaluate} whether delegated work was actually accomplished. Every child is a fresh instance of the parent, so all three functions live in a single policy. Training the agent well means giving each capability its own signal without the three interfering on shared parameters.

A self-invented subtask has no ground-truth check, so the agent must judge for itself whether it succeeded. We make this self-evaluation concrete with a \emph{rubric}: when the parent delegates a subtask, it commits a short set of weighted natural-language success criteria for the child before the child runs, so they cannot be fitted to the trajectory they will later score. Applying the rubric to the child's output yields a continuous, subtask-specific score in place of the coarse per-node signal.


However, a rubric helps only when its criteria align with subtask success. An untrained policy writes misaligned criteria, and using its scores as rewards helps little, inconsistently across tasks (\S\ref{sec:experiments}). We therefore train rubric generation, but the signal for that training cannot be the task reward itself: since rubric generation and solving share weights, rewarding the rubric through task reward would let the two co-adapt, inflating rubric scores together without improvement. Instead, the signal comes from an external check: we score subtask rollouts with a verifier or judge and train the rubric to rank verified successes above failures. Importantly, the judge is used only here, never to reward nodes directly as common approaches do; the rubric provides the dense per-node signal itself, so external supervision is lighter overall. Grounded this way, shared weights become an asset: learning to judge subtasks reshapes how the policy solves them, which is our central empirical finding (\S\ref{sec:experiments}).


A capability, \emph{subtask decomposition}, is the choice of how to split a task into subproblems to delegate. We assign it a reward. In a solved tree, a delegation is credited by the fraction of the task's leaf-level work its subtree covers, so a delegation that shoulders a large, well-scoped share of the solution scores high, while trivial hand-offs score near zero and delegations in failed trees score zero. This rewards the agent for using delegation to carry structural load, something no execution signal captures. The reward adds little on shallow, near-independent subtasks but helps when subtasks are interdependent, as in tasks with dependency chains, shared state, or coordination (\S\ref{sec:experiments}).

Finally, since the learned rubric captures genuine subtask quality, it is useful beyond training. At inference we search the tree depth-first and select among candidate resolutions by rubric score, propagating the best upward. This further improves final task success without additional training (\S\ref{sec:tts}). The evaluation capability thus pays off as a reusable inference-time selector, a further scaling axis for recursive agents. We call the method \textbf{SERA} (Self-Evaluating Recursive Agents). Our contributions are as follows:

\begin{enumerate}[leftmargin=*, itemsep=0pt, topsep=0pt]
\item A framing of recursive-agent training as learning three coupled capabilities (decomposing, solving, and evaluating subtasks) in a single shared policy, with evidence that each requires its own training signal (\S\ref{sec:method}).\looseness-1
\item A learned self-evaluation mechanism: policy-generated rubrics trained by a ranking objective against verified outcomes. We show that training rubric generation is what improves execution, that it reduces external supervision while outperforming direct use of the judge (\S\ref{sec:method}).
\item A leaf-coverage decomposition reward that credits a delegation by how much of the verified task it covers, with gains concentrated when subtasks are interdependent (\S\ref{sec:method}).
\item State-of-the-art results on TextCraft-Synth and TextWorld-Sync (\S\ref{sec:experiments}), and the learned rubric as an inference-time selector for tree search, with further gains (\S\ref{sec:tts}).
\end{enumerate}

\vspace{-5pt}
\section{Related Work}
\label{sec:related}
\vspace{-3pt}

\textbf{Recursive and multi-agent training.~}
Recursive and multi-agent scaffolds decompose tasks and route subproblems to child agents, scaling language-model agents at inference time \citep{prasad2024adapt, schroeder2025thread}, but most treat recursion as a fixed scaffold around a frozen model. Recent work instead trains agents to delegate, though largely at depth~1 with fixed or restricted workers: parallel delegation with reinforcement learning \citep{team2026kimi} and jointly trained organizer--worker systems \citep{sun2025scaling, chi2025era}. Recursive Agent Optimization (RAO) is the closest prior work, combining recursion beyond depth~1, asynchronous subagent execution, and end-to-end RL training of both root agents and subagents \citep{gandhi2026recursive}. We adopt this setting and use RAO as our primary baseline, but depart from it in the signal each node receives: RAO scores nodes with a binary success label from a verifier or judge and says nothing about decomposition quality, whereas SERA gives each node a trained continuous rubric and credits decomposition directly.\looseness-1

\textbf{Dense rewards and process supervision for agents.~}
A large body of work densifies sparse outcome rewards with step- or process-level supervision, from process reward models \citep{lightman2024let} to agentic variants that assign credit to intermediate actions \citep{zhang2026trca, li2026and}; these operate on flat, linear trajectories with the intermediate signal supplied by a separate critic or an additive decomposition of the return. Classical hierarchical RL such as options \citep{dietterich2000hierarchical} and goal-conditioned hierarchies \citep{nachum2018data} assigns credit across levels but assumes a fixed set of subtask types or hierarchy depth. In contrast, SERA operates on a dynamically generated tree where the hierarchy, the subtasks, and the evaluator are all produced by a single policy and vary from rollout to rollout.

\textbf{Rubric-based reinforcement learning.~}
A separate question from where credit is assigned is what form the per-node signal takes. Rubrics extend RL beyond exactly verifiable domains by scoring outputs against generated criteria \citep{liu2026openrubrics}. Because static rubrics saturate as the policy improves, recent work co-trains the rubric generator with the policy through per-step return decomposition \citep{tian2026arco}, alternating generator--judge optimization \citep{xu2026alternating}, or adversarial co-evolution \citep{ding2026evorubrics}, with RubricEM the closest in domain family \citep{RubricEM}. All assume flat or fixed-stage computation with a rubric model separate from the policy. SERA instead applies learned rubric generation to recursive, policy-invented subtasks whose topology varies across rollouts, uses the policy itself as the evaluator, and trains rubrics with a ranking objective anchored to verified outcomes. We find that the label-free discriminativeness signal favored in prior rubric learning degrades performance in a tree, where score spread reflects subtask-difficulty variation more than rubric quality (\S\ref{sec:ablation-study}); relatedly, work documenting reward hacking under learned reward signals \citep{gao2025scaling, wang2026reproducing, yang2026rubric} motivates our verified-outcome anchoring.\looseness-1

\begin{figure}[t]
    \centering
    \includegraphics[width=0.86\textwidth]{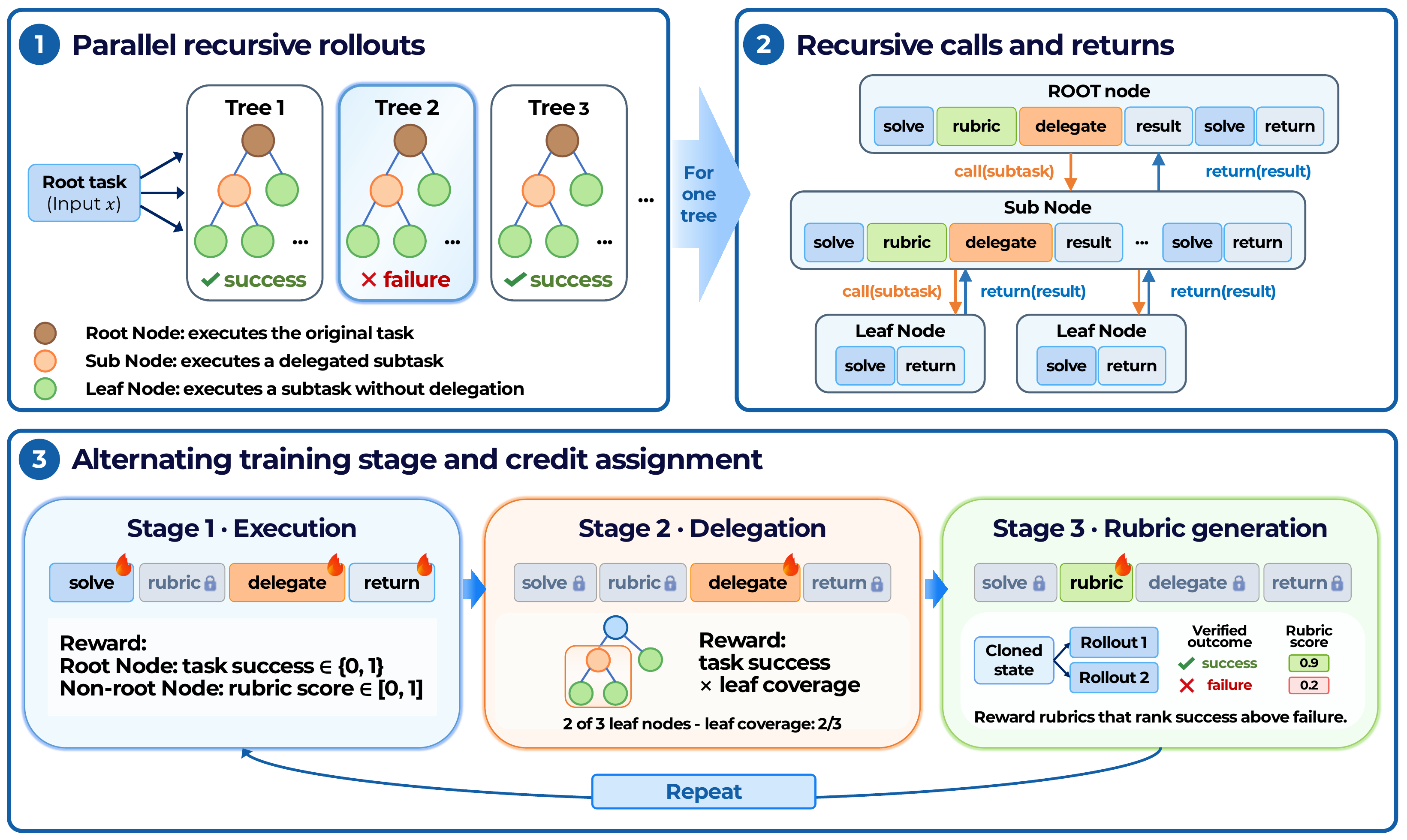}
    \caption{Recursive Agent Optimization with rubric-based trajectory
    scoring. A shared-policy root agent delegates subtasks recursively; each completed SubAgent trajectory can receive a Rubric Reward together with the Program or LLM Judge reward.}
    \label{fig:sera_ppl}
    \vspace{-8pt}
\end{figure}

\vspace{-7pt}
\section{Method}
\label{sec:method}
\vspace{-3pt}

\vspace{-4pt}
\subsection{Problem setup}
\vspace{-3pt}

SERA trains a single policy $\pi_\theta$ to decompose a root task, execute the resulting subtasks, and evaluate their outcomes (Figure~\ref{fig:sera_ppl}). Given a root task $x$, the policy either solves it directly or emits a \emph{delegation action} that creates a child agent for a self-generated subtask. Recursively applying the same policy produces a tree $T$, where each node $v\in T$ has a subtask $X_v$, trajectory $\tau_v$, output $o_v$, final environment state $z_v$, child set $C(v)$, and depth $d(v)$. The root task is evaluated by the environment, yielding $R\in[0,1]$, whereas intermediate subtasks receive self-evaluated rewards during execution. For variance reduction, we sample $K>1$ independent trees $\{T_i\}_{i=1}^K$ for each root task.\looseness-1

We assign a separate learning signal to each capability. The \emph{execution reward} gives every node a continuous score from a policy-generated rubric, training the agent to solve subtasks well. The \emph{leaf-coverage reward} credits each delegation decision by the share of the solved task it organizes, training the agent to decompose effectively. The \emph{rubric-generation reward} trains the agent to write evaluation criteria that track verified outcomes, using a ranking objective over counterfactual rollouts. These three signals are optimized by alternating updates on the shared policy. To prevent evaluation from adapting to the outcome it scores, each rubric obeys an \emph{ex-ante commitment}: once generated by the parent, it is frozen before the child's trajectory and final state are observed.

\vspace{-5pt}
\subsection{Self-evaluated execution reward}
\label{sec:rubric-execution}
\vspace{-3pt}

When a parent delegates a subtask, it generates a frozen, weighted rubric
\begin{equation}\textstyle
  \rho_v=\{(k_{v,j},w_{v,j})\}_{j=1}^{J_v},
  \qquad w_{v,j}\geq 0,\qquad \sum_{j=1}^{J_v}w_{v,j}=1,
\end{equation}
where $k_{v,j}$ is a natural-language success criterion and $w_{v,j}$ is its weight. The generator conditions only on information available at delegation time and never observes the child's future trajectory or output.\looseness-1

After the child finishes, a separate invocation of the same policy (run without gradient) scores the frozen rubric against the child's trajectory and final state. The scorer produces a structured output with two components: a binary success gate $g_v\in\{0,1\}$ indicating whether the child completed the delegated subtask, and a per-criterion score $c_{v,j}\in[0,1]$ for each criterion $k_{v,j}$. If the child did not complete the subtask (e.g., it exhausted its step budget or produced no usable output), the gate enforces $g_v=0$ and the score is exactly zero, since grading individual criteria on an incomplete attempt would introduce noise. The continuous node score is $s_v = g_v\sum_{j=1}^{J_v}w_{v,j}c_{v,j}\in[0,1]$, and the node-level execution reward assigns $r^{\mathrm{exec}}_{i,v}=R_i$ at the root of $T_i$ and $r^{\mathrm{exec}}_{i,v}=s_{i,v}$ at every other node. During the execution phase, rubric generation and scoring do not receive gradient; only the execution trajectory is updated. Rubric generation is trained separately through the ranking objective described in Section~\ref{sec:rubric-training}.

\vspace{-5pt}
\subsection{Leaf-coverage reward for decomposition}
\label{sec:leaf-reward}
\vspace{-3pt}

Execution credit evaluates what an agent does after a decision, but it does not directly evaluate whether the decision to delegate was structurally useful. We therefore assign a separate reward to each delegation event. We measure coverage by leaf nodes (nodes with no children), since leaves are the nodes that solve atomic subproblems and contribute directly to the final answer, while internal nodes serve as orchestrators. Leaf coverage thus captures how much of the final solution is produced through a given delegation, independent of tree depth. Since no additional signal is available to measure each leaf's individual contribution, we weight all leaves equally; incorporating subtask-level importance weights is a natural extension. Let $e$ be a delegation event in tree $T_i$, let $L_i(e)$ be the leaves in the subtree created by $e$, and let $L(T_i)$ be the leaves of the complete root-task tree. The leaf coverage and delegation reward are
\begin{equation}\textstyle
  q^{\mathrm{leaf}}_{i,e}
  =\frac{|L_i(e)|}{|L(T_i)|},
  \qquad
  r^{\mathrm{leaf}}_{i,e}
  =R_i\,q^{\mathrm{leaf}}_{i,e}.
\end{equation}
Thus, a delegation receives credit only when the complete root task succeeds, and its credit is proportional to the amount of leaf-level work organized by its subtree. All delegation events in a failed tree receive zero. The reward is applied only to the tokens that specify the delegation goal; neither the associated execution tokens nor the rubric tokens receive this reward. It consequently supplies a direct optimization signal for decomposition while remaining complementary to the execution reward.\looseness-1

\vspace{-5pt}
\subsection{Training rubric generation from verified outcomes}
\label{sec:rubric-training}
\vspace{-3pt}

Rubric scores are useful only if the rubric criteria correlate with actual subtask success. The most readily available signal for training rubric generation is the root task reward, but because the rubric generator and the solver share $\pi_\theta$, optimizing rubrics through this reward would allow the two to co-adapt, inflating rubric scores without improving actual quality. We instead train rubric generation against verified subtask outcomes obtained from counterfactual rollouts. Specifically, we use a ranking loss that requires the rubric to score verified successes above verified failures on the same subtask. This isolates rubric quality from absolute task difficulty, since the rubric need only separate outcomes within a single subtask rather than calibrate scores across different ones.

Concretely, during rubric-generation training we clone the environment state at each delegation point and sample a group of $B$ rollouts of the same subtask. One rollout continues the original recursive execution, while the remaining $B-1$ rollouts serve solely as counterfactual training data for the rubric generator. Note that this cloning and counterfactual sampling is required only for rubric-generation training; the execution and decomposition phases operate on ordinary rollouts without additional sampling. Each rollout $b$ receives a verified outcome $y_b\in[0,1]$ from the environment verifier or an external judge, and the frozen rubric assigns it a score $s_b$. We form all ordered pairs in which one rollout has a strictly higher verified outcome than another, and define the margin-ranking loss\looseness-1
\begin{equation}\textstyle
  \mathcal{L}^{\mathrm{rank}}
  =\frac{1}{|\mathcal{P}|}
    \sum_{(i,j)\in\mathcal{P}}
    \max\!\bigl(0,\;\delta-s_i+s_j\bigr),
  \qquad
  r^{\mathrm{rub}}=-\mathcal{L}^{\mathrm{rank}},
\end{equation}
where $\mathcal{P}=\{(i,j):y_i>y_j\}$ and $\delta>0$ is the ranking margin. The reward is high when the rubric ranks rollouts in the same order as the verified outcomes, with a margin of at least $\delta$. Groups in which all rollouts receive the same verified outcome provide no pairwise information and are skipped. The verified outcomes are used only for this rubric-generation objective; during ordinary execution, the policy scores a child using its own frozen rubric.

\vspace{-5pt}
\subsection{Advantage estimation and policy objective}
\label{sec:method-objective}
\vspace{-3pt}

The preceding subsections defined three raw rewards: an execution reward for solving, a leaf-coverage reward for decomposition, and a ranking reward for rubric generation. To form a policy gradient, each raw reward is centered by a leave-one-out (LOO) baseline to produce an advantage. Since the three rewards operate at different granularities, each uses its own baseline group. We define the three baselines below and then state the shared policy objective.

\textbf{Execution advantage.~} For execution, every node in tree $T_i$ uses the root outcomes of the other independently sampled trees as its baseline:
\begin{equation}\textstyle
  A^{\mathrm{exec}}_{i,v}
  =r^{\mathrm{exec}}_{i,v}
   -\frac{1}{K-1}\sum_{j\neq i}R_j.
\end{equation}
The baseline is independent of the trajectory in $T_i$, so an intermediate node can receive positive execution credit when its own rubric score is high even if the complete tree fails. A tighter baseline would average over multiple rollouts of the \emph{same} subtask, but because each subtask is policy-generated and varies across trees, no matched comparison group exists at depth $\geq 1$. The root-group baseline is a practical compromise that remains unbiased while using the available between-tree variation.\looseness-1

\textbf{Decomposition advantage.~} For a tree $T_i$, let $E_i$ be its delegation events and define its mean leaf reward $\bar r^{\mathrm{leaf}}_i=\frac{1}{|E_i|}\sum_{e\in E_i}r^{\mathrm{leaf}}_{i,e}$, with trees that contain no delegation events omitted from the leaf update. The leaf advantage for event $e$ is
\begin{equation}\textstyle
  A^{\mathrm{leaf}}_{i,e}
  =r^{\mathrm{leaf}}_{i,e}
   -\frac{1}{K-1}\sum_{j\neq i}\bar r^{\mathrm{leaf}}_j.
\end{equation}
Averaging per tree before computing the LOO baseline ensures that each tree contributes equally regardless of how many delegation events it contains; without this, trees with more delegations would dominate the baseline and bias the advantage toward shallower or deeper decomposition patterns.\looseness-1

\textbf{Rubric-generation advantage.~} For rubric generation, let $r^{\mathrm{rub}}_m$ be the ranking reward for the $m$-th delegation event in the batch whose counterfactual group contains both successes and failures (all-success or all-failure groups are skipped since they contain no ordered pairs to rank). Writing $M$ for the number of such valid events, the advantage is
\begin{equation}\textstyle
  A^{\mathrm{rub}}_m
  =r^{\mathrm{rub}}_m
   -\frac{1}{M-1}\sum_{m'\neq m}r^{\mathrm{rub}}_{m'}.
\end{equation}
\textbf{Shared policy objective.~} All three advantages are optimized with the same loss, CISPO \citep{minimax2025m1}, a clipped importance-sampling policy gradient. For a token $a_t$ with context $h_t$ and advantage $\tilde{A}_t$ computed from whichever reward is being optimized in the current step, the per-token loss is\looseness-1
\begin{equation}\textstyle
  \ell_t = -\operatorname{sg}\!\bigl(\operatorname{clip}(\varrho_t,\, c_{\min},\, c_{\max})\bigr)\,\tilde{A}_t\,\log\pi_\theta(a_t\mid h_t), \qquad \varrho_t = \frac{\pi_\theta(a_t\mid h_t)}{\pi_{\mathrm{old}}(a_t\mid h_t)},
\end{equation}
where $\operatorname{sg}$ denotes stop-gradient. Let $\mathcal{T}$ be the set of tokens supervised by the active reward, and let $\omega_t \propto 1/N_{d(t)}$ be an inverse-frequency depth weight, where $N_{d(t)}$ is the number of tokens at the depth of token $t$. The total loss is
\begin{equation}\textstyle
  \mathcal{L}_{\mathrm{CISPO}} = \frac{1}{\sum_{t\in\mathcal{T}}\omega_t}\sum_{t\in\mathcal{T}}\omega_t\,\ell_t.
\end{equation}
This ensures that each recursion depth contributes equally to the gradient regardless of how many nodes it contains \citep{gandhi2026recursive}. The three updates share this objective, differing only in their raw reward, the LOO baseline group, and the set $\mathcal{T}$ of tokens they supervise.

\vspace{-7pt}
\subsection{Alternating optimization schedule}
\label{sec:training-schedule}
\vspace{-4pt}

SERA jointly optimizes three objectives on shared parameters. Combining all three gradients in a single update risks instability, since the objectives can interfere and their scales differ. We instead alternate: each optimizer step activates exactly one objective, and the updates cycle in the fixed ratio $n_{\mathrm{exec}}:n_{\mathrm{leaf}}:n_{\mathrm{rub}}$. Execution updates receive the largest share because, in preliminary experiments, execution training is the primary driver of performance and covers the majority of tokens in a typical tree. The leaf-coverage and rubric-generation stages are lighter, each requiring only a small number of periodic updates to improve decomposition and refine the rubric that supplies the dense execution signal. Alternation keeps the credit sources well separated, and the ratio can be adjusted per domain without changing the method. The same learned rubric is then reused at inference time to rank candidate node resolutions during the bottom-up tree-search procedure described in Section~\ref{sec:tts}.

\vspace{-7pt}
\section{Experiments}
\label{sec:experiments}
\vspace{-5pt}


\textbf{Benchmarks.~}
We evaluate on two benchmarks that differ in a property central to our method: whether intermediate subtasks can be verified independently.
\textit{TextCraft-Synth}~\citep{gandhi2026recursive} is a procedural crafting environment requiring hierarchical planning, tool use, and recursive delegation, in which the agent crafts a target item by looking up recipes and delegating sub-crafts to child agents. Crucially, each subtask has a deterministic program verifier returning a binary pass/fail outcome, so SERA's rubric training requires no external judge here. Tasks span four difficulty levels defined by recipe depth; we train on Medium and evaluate across all levels to test generalization.
\textit{TextWorld-Sync} is our multi-agent extension of TextWorldExpress~\citep{jansen2023textworldexpress}, a shared-state text environment for evaluating coordination under constraints, with seven task families spanning dependency chains, resource conflicts, load balancing, and hierarchical decomposition. Unlike TextCraft, subtask outcomes are not independently verifiable; rubric training instead uses an LLM judge that provides binary success/failure labels for counterfactual rollouts, making this the benchmark that tests SERA's ability to amortize a sparse external signal. On both benchmarks the verified outcome $y_b$ is binary, so the ranking loss and leaf-coverage reward from Section~\ref{sec:method} apply without modification. Detailed dataset statistics and task-family descriptions are in Appendix~\ref{apd:text_world_dataset}.\looseness-1

\textbf{Baselines.~}
We compare against three types of baselines: prompting, single-agent RL, and recursive multi-agent RL.
(1)~\textit{Prompting}: Qwen3-4B-Instruct in single-agent and multi-agent (recursive delegation enabled) modes. (2)~\textit{Single-agent RL}: GRPO~\citep{shao2024deepseekmath} and CISPO~\citep{minimax2025m1}, both trained without delegation. (3)~\textit{Recursive multi-agent RL}: RAO~\citep{gandhi2026recursive}, which jointly trains root and subagent trajectories with per-node binary rewards. 
To isolate each component's contribution, we also evaluate ablations that add one capability to RAO (\textit{RAO + Decomposition}, \textit{RAO + Rubric Training}) or remove one or more from the full method (\textit{\method{} w/o Decomposition} and \textit{\method{} w/o Decomposition \& Rubric Training}); the exact reward composition of each is given in Appendix Table~\ref{tab:training_method_schedule}. All reported success rates are determined by deterministic environment verification of the root task.

\textbf{Training details.~}
All methods use Qwen3-4B-Instruct-2507~\citep{yang2025qwen3} as the base model, fully fine-tuned with Adam. All compared recursive methods share the advantage-estimation and policy-optimization pipeline described in Section~\ref{sec:method-objective}, differing only in their reward and credit assignments; \method{}'s alternating schedule uses the ratio $n_{\mathrm{exec}}:n_{\mathrm{leaf}}:n_{\mathrm{rub}}=16:2:2$. Each agent executes at most 20 environment steps at recursion depth up to 3. We report TextCraft-Synth at 250 optimizer updates and TextWorld-Sync at 400; remaining hyperparameters are listed in Appendix Table~\ref{tab:shared_training_settings}.\looseness-1

\vspace{-5pt}
\subsection{Experimental Results}
\label{sec:experimental-results}
\vspace{-3pt}

\begin{table}[t]
\setcitestyle{numbers,square}
\centering
\caption{Main results on TextCraft-Synth and TextWorld-Sync, grouped into single- and multi-agent methods. Success rates (\%) are per difficulty level; the shaded \emph{Avg.} column is the task-count-weighted mean over the four levels, averaged over 3 seeds ($\pm$ std). \textbf{Bold}/\underline{underline}: best/second-best per column. D\,=\,Decomposition reward, RT\,=\,Rubric Training.}
\label{tab:main_compare}
\setlength{\tabcolsep}{3.5pt}
\resizebox{\textwidth}{!}{
\begin{tabular}{lcccc>{\columncolor{avgcol}}c|cccc>{\columncolor{avgcol}}c}
\toprule
\multirow{2}{*}{Method}
& \multicolumn{5}{c|}{\textbf{TextCraft}}
& \multicolumn{5}{c}{\textbf{TextWorld}} \\
\cmidrule(lr){2-6} \cmidrule(lr){7-11}
& Easy & Medium & Hard & Extreme & Avg.
& Easy & Medium & Hard & Extreme & Avg. \\
\midrule
\multicolumn{11}{@{}l}{\emph{Single-Agent}} \\
Qwen3-4B-Instruct
& 79.37$\pm$0.79 & 13.77$\pm$1.78 & 0.00$\pm$0.00 & 0.00$\pm$0.00 & 23.10$\pm$0.42
& 26.57$\pm$0.29 & 25.43$\pm$1.12 & 0.00$\pm$0.00 & 0.00$\pm$0.00 & 13.00$\pm$0.29 \\
CISPO
& 94.56$\pm$0.30 & 32.86$\pm$0.62 & 0.00$\pm$0.00 & 0.00$\pm$0.00 & 33.07$\pm$0.22
& \underline{99.14}$\pm$0.00 & 56.29$\pm$1.41 & 0.00$\pm$0.00 & 0.00$\pm$0.00 & 38.86$\pm$0.35 \\
GRPO
& 90.48$\pm$0.64 & 23.47$\pm$0.93 & 0.00$\pm$0.00 & 0.00$\pm$0.00 & 28.96$\pm$0.35
& \underline{99.14}$\pm$0.00 & 58.00$\pm$2.68 & 0.00$\pm$0.00 & 0.00$\pm$0.00 & 39.29$\pm$0.67 \\
\midrule
\multicolumn{11}{@{}l}{\emph{Multi-Agent}} \\
Qwen3-4B-Instruct
& 74.15$\pm$0.68 & 26.76$\pm$1.69 & 0.49$\pm$0.85 & 0.00$\pm$0.00 & 26.37$\pm$0.64
& 54.19$\pm$0.59 & 12.95$\pm$0.33 & 0.00$\pm$0.00 & 0.00$\pm$0.00 & 16.79$\pm$0.21 \\
RAO
& \textbf{99.09}$\pm$0.39 & 89.05$\pm$0.27 & 70.34$\pm$3.77 & 3.43$\pm$1.12 & 68.93$\pm$0.78
& 98.95$\pm$1.08 & 85.62$\pm$2.29 & 20.19$\pm$3.00 & 2.95$\pm$2.16 & 51.93$\pm$1.42 \\
RAO + D
& 98.19$\pm$0.39 & 96.24$\pm$0.47 & 60.54$\pm$4.31 & 2.94$\pm$1.47 & 68.93$\pm$1.17
& 97.71$\pm$0.29 & \underline{89.33}$\pm$2.31 & 16.76$\pm$0.87 & 1.52$\pm$0.16 & 51.33$\pm$0.50 \\
RAO + RT
& 98.19$\pm$1.04 & 93.58$\pm$0.27 & 76.96$\pm$1.12 & \underline{7.11}$\pm$1.53 & 72.47$\pm$0.16
& \textbf{99.52}$\pm$0.33 & 77.90$\pm$0.72 & 5.05$\pm$0.87 & 0.00$\pm$0.00 & 45.62$\pm$0.46 \\
\cmidrule(lr){1-11}
\method{} w/o D\,\&\,RT
& \underline{98.87}$\pm$1.04 & \underline{96.87}$\pm$0.27 & 73.28$\pm$1.53 & 3.92$\pm$1.70 & 72.26$\pm$0.18
& 92.67$\pm$1.32 & 87.14$\pm$1.43 & 15.24$\pm$0.44 & \underline{9.33}$\pm$1.19 & 51.10$\pm$0.15 \\
\method{} w/o D
& 98.41$\pm$1.04 & 95.77$\pm$0.81 & \textbf{79.90}$\pm$2.78 & \textbf{11.52}$\pm$2.25 & \textbf{74.84}$\pm$1.41
& 97.33$\pm$0.59 & \textbf{92.57}$\pm$1.48 & \underline{45.71}$\pm$2.54 & 5.43$\pm$1.78 & \underline{60.26}$\pm$1.16 \\
\method{}
& \textbf{99.09}$\pm$0.79 & \textbf{97.81}$\pm$1.43 & \underline{78.92}$\pm$4.05 & 6.13$\pm$0.85 & \underline{74.31}$\pm$0.90
& 98.19$\pm$0.87 & 84.95$\pm$2.86 & \textbf{54.29}$\pm$7.78 & \textbf{22.86}$\pm$4.23 & \textbf{65.07}$\pm$3.48 \\
\bottomrule
\end{tabular}
}
\end{table}

Table~\ref{tab:main_compare} compares prompting-based agents, single-agent RL, and recursive multi-agent RL methods on TextCraft-Synth and TextWorld-Sync. We highlight key findings below.

\textbf{\method{} achieves the highest or second-highest average success on both benchmarks}: $74.31\%$ on TextCraft-Synth (second) and $65.07\%$ on TextWorld-Sync (first), improving over RAO by $5.38$ and $13.14$ points respectively. \method{} differs from RAO in three ways: continuous rubric-based execution rewards, trained rubric generation, and a decomposition reward. We isolate each below.

\textbf{Recursive delegation must be explicitly trained.} The prompted Multi-Agent baseline improves over its Single-Agent counterpart by only $3.27$ points on TextCraft-Synth and $3.79$ points on TextWorld-Sync. In contrast, RAO substantially outperforms Single-Agent GRPO by $39.97$ and $12.64$ points, and Single-Agent CISPO by $35.86$ and $13.07$ points, on the two benchmarks respectively. Merely exposing the model to a delegation interface is insufficient; the policy must be explicitly optimized to decide when to delegate, what to assign, and how to integrate the result.

\textbf{Rubric rewards without rubric training do not reliably improve execution.} \method{} w/o D\,\&\,RT replaces binary per-node rewards with continuous rubric scores from an untrained generator. Despite the denser signal, it improves average success by only $3.33$ points on TextCraft-Synth and underperforms RAO by $0.83$ points on TextWorld-Sync.

\textbf{Training rubric generation drives the largest gains.} Compared with \method{} w/o D\,\&\,RT, which uses rubric-based execution rewards but no rubric training, \method{} w/o D additionally trains rubric generation. This single change improves average success by $2.58$ points on TextCraft-Synth and $9.16$ points on TextWorld-Sync. The gains are particularly pronounced on difficult tasks: Hard and Extreme success on TextCraft-Synth improve by $6.62$ and $7.60$ points respectively.

\textbf{Rubric training requires rubric-based execution rewards.} Conversely, RAO + RT adds rubric-generation training to RAO but keeps binary execution rewards. It is competitive on TextCraft-Synth ($72.47\%$ avg.) but collapses to $45.62\%$ on TextWorld-Sync, well below RAO ($51.93\%$). Without rubric-based execution rewards the trained rubric never feeds back into how subtasks are solved; the two are thus coupled, and training helps only when the trained rubric also serves as the execution signal.\looseness-1

\textbf{Decomposition training helps most on difficult tasks.}
\method{} exceeds \method{} w/o D by $4.81$ points on TextWorld-Sync, with the gains concentrated on Hard ($+8.58$) and Extreme ($+17.43$) tasks. On TextCraft-Synth, \method{} trails \method{} w/o D by $0.53$ points. The decomposition reward yields gains primarily in combination with rubric-based execution rewards: added to RAO alone it changes average success by only $0.00$ and $-0.60$ points on the two benchmarks.

\begin{wraptable}{r}{0.52\textwidth}
\centering
\scriptsize
\setlength{\tabcolsep}{3pt}
\vspace{-18pt}
\caption{External-supervision cost accumulated over training. On TextCraft-Synth, subtask outcomes come from a free program verifier, so we report only its call count; on TextWorld-Sync, from an external LLM judge, so we also report input/output token usage. Calls in thousands (K), tokens in millions (M).}
\label{tab:external_model_tokens}
\vspace{2pt}
\resizebox{\linewidth}{!}{%
\begin{tabular}{lc|ccc}
\toprule
\multirow{2}{*}{Method}
& \textbf{TextCraft}
& \multicolumn{3}{c}{\textbf{TextWorld}} \\
\cmidrule(lr){2-2} \cmidrule(lr){3-5}
& Verifier (K)
& Judge Calls (K) & Input (M) & Output (M) \\
\midrule
RAO & 550.94 & 147.80 & 633.80 & 16.13 \\
RAO + D & 381.65 & 84.22 & 371.20 & 9.10 \\
RAO + RT & 769.02 & 251.46 & 847.18 & 27.07 \\
\cmidrule(lr){1-5}
\method{} w/o D\,\&\,RT & 20.24 & 0.00 & 0.00 & 0.00 \\
\method{} w/o D & 366.52 & 43.70 & 136.35 & 5.34 \\
\method{} & 257.96 & 35.79 & 125.32 & 4.27 \\
\bottomrule
\end{tabular}%
}
\vspace{-2pt}
\end{wraptable}

\textbf{SERA uses external supervision to train a reusable evaluator.}
Table~\ref{tab:external_model_tokens} shows that \method{} reduces TextCraft's program-verifier calls from $550.94$K to $257.96$K, approximately $2.1\times$. The saving is more consequential on TextWorld, where intermediate supervision requires an external LLM: calls decrease from $147.80$K to $35.79$K ($4.1\times$), and input tokens from $633.80$M to $125.32$M ($5.1\times$). SERA concentrates these labels on rubric-generation training, allowing the learned evaluator to supply subsequent execution rewards locally. Removing rubric training eliminates external LLM calls entirely, but \method{} w/o D\,\&\,RT reaches only $51.10\%$ on TextWorld, versus SERA's $65.07\%$. The useful trade-off is therefore selective external grounding coupled with learned self-evaluation.

\vspace{-5pt}
\subsection{Ablation Study}
\label{sec:ablation-study}
\vspace{-3pt}

Table~\ref{tab:rubric_ablation} ablates two aspects of the rubric mechanism on TextCraft-Synth: how the rubric is constructed and scored (top block), and how rubric generation is trained (bottom block).

\textbf{Subtask-specific rubrics enable competitive self-evaluation.}
Here, \emph{Generator} identifies the model that writes the criteria, while \emph{Scorer} identifies the model that applies them to completed trajectories. Without explicit rubric-generation training, using the policy as both the generator and the scorer of a fresh per-subtask rubric achieves $72.26\%$ average success, compared with $71.41\%$ when the external frontier model Kimi is the generator (policy scorer) and $73.10\%$ when Kimi is both generator and scorer. Even in this training-free regime, self-generated rubrics are competitive with the external judge (within $0.84$ points, and within seed noise), so keeping evaluation in-policy costs essentially nothing before training. By contrast, replacing the per-subtask generator with a single global policy rubric reused across all subtasks reduces success to $67.77\%$ (policy scorer). This $4.49$-point gap suggests that adapting criteria to the delegated task is more useful than reusing a single rubric across subtasks.\looseness-1

\textbf{Effective rubric training requires alignment, not merely score dispersion.}
Training rubric generation against verified outcomes raises average success from $72.26\%$ to $74.84\%$, outperforming every variant without explicit rubric training. Notably, this trained policy self-evaluator also surpasses the strongest training-free configuration, using Kimi as both generator and scorer ($73.10\%$), by $1.74$ points, with the margin concentrated on the hardest tasks (Hard: $79.90$ vs.\ $77.45$; Extreme: $11.52$ vs.\ $4.17$). A trained $4$B policy scoring its own rubrics thus supplies better execution rewards than a frontier external judge, so effective self-evaluation does not require an external model. We further test a variance-based adaptation of the discriminativeness component in \citet{ding2026evorubrics}, which rewards a rubric for spreading scores across a group of candidate rollouts. Introducing this signal reduces success to $65.77\%$, a $9.07$-point drop. Score variance alone does not indicate whether higher scores correspond to successful execution: it is unchanged when a group's ranking is reversed and can increase by separating equally unsuccessful rollouts. Such distinctions can provide misleading execution rewards when evaluation and solving share a policy. The degradation is consistent with this mismatch and supports grounding rubric training in verified success--failure comparisons rather than rewarding dispersion itself.

\begin{table}[t]
\setcitestyle{numbers,square}
\centering
\scriptsize
\caption{Ablation of rubric design choices on TextCraft-Synth. All rows use rubric scores as the execution reward and differ only in three axes: the \emph{Generator} that writes the rubric (the SERA \emph{Policy}, the external frontier \emph{Kimi} judge, or a single \emph{global} policy rubric reused across subtasks---otherwise per-subtask), the \emph{Scorer} that applies it, and the \emph{Training} of rubric generation (``--'' none; \emph{Rank} our verified-outcome ranking objective; \emph{Rank+Disc.} adds the label-free discriminativeness reward of \citet{ding2026evorubrics}). Success rates (\%) by difficulty with overall average, averaged over 3 seeds. \textbf{Bold}/\underline{underline}: best/second-best per column.}
\label{tab:rubric_ablation}
\setlength{\tabcolsep}{5pt}
\begin{tabular}{lllcccc>{\columncolor{avgcol}}c}
\toprule
Generator & Scorer & Training & Easy & Medium & Hard & Extreme & Avg. \\
\midrule
\multicolumn{8}{@{}l}{\emph{Training-free rubric generation}} \\
Policy            & Policy & --        & \underline{98.87}$\pm$1.04 & 96.87$\pm$0.27 & 73.28$\pm$1.53 & 3.92$\pm$1.70 & 72.26$\pm$0.18 \\
Kimi              & Policy & --        & \underline{98.87}$\pm$1.04 & \textbf{97.97}$\pm$0.27 & 68.38$\pm$3.21 & 3.19$\pm$1.12 & 71.41$\pm$0.37 \\
Kimi              & Kimi   & --        & 97.73$\pm$0.39 & \underline{97.34}$\pm$1.51 & \underline{77.45}$\pm$1.70 & \underline{4.17}$\pm$1.70 & \underline{73.10}$\pm$1.11 \\
Policy \emph{(global)} & Policy & --   & \textbf{99.09}$\pm$1.04 & 91.55$\pm$2.82 & 62.01$\pm$2.58 & 2.45$\pm$0.42 & 67.77$\pm$1.14 \\
\midrule
\multicolumn{8}{@{}l}{\emph{Trained rubric generation}} \\
Policy            & Policy & Rank      & 98.41$\pm$1.04 & 95.77$\pm$0.81 & \textbf{79.90}$\pm$2.78 & \textbf{11.52}$\pm$2.25 & \textbf{74.84}$\pm$1.41 \\
Policy            & Policy & Rank+Disc.& 97.28$\pm$0.68 & 92.64$\pm$0.27 & 52.45$\pm$5.00 & 2.94$\pm$0.74 & 65.77$\pm$1.21 \\
\bottomrule
\end{tabular}
\end{table}

\vspace{-5pt}
\subsection{Test-time Scaling with Learned Self-Evaluation}
\label{sec:tts}
\vspace{-3pt}

We test whether the evaluation capability learned during training can guide additional computation at inference. We compare Qwen3-4B-Instruct without SERA training against the SERA-trained policy on TextWorld-Sync. With $N=2$, we fork an additional candidate branch at each selection point and retain the higher-scoring resolution, using either the current policy (\emph{Policy judge}) or an external Kimi-K3 model~\citep{team2026kimi-k3} (\emph{Kimi judge}) to apply the rubric. Selection proceeds bottom-up, with child results fixed before selecting the parent resolution. The $N=1$ setting uses a single rollout without branching or selection.

\textbf{Self-evaluation yields further gains without additional training.}
As shown in Table~\ref{tab:test_time_scaling}, policy-guided selection raises \method{}'s average success from $65.07\%$ to $67.50\%$ ($+2.43$ points), with Extreme-task success increasing from $22.86\%$ to $28.95\%$. The trained policy judge is on par with the external Kimi-K3 judge, with nearly identical averages ($67.50\%$ vs.\ $67.29\%$) and comparable success on the hardest levels (Hard $55.71\%$ vs.\ $55.52\%$, Extreme $28.95\%$ vs.\ $27.62\%$). Thus, the learned evaluator supports further improvement through inference-time selection without an external judge, extending its role beyond supplying training rewards.

\textbf{SERA training increases the benefit of test-time selection.}
For Qwen3-4B-Instruct without SERA training, the policy judge yields no average gain ($-0.22$ points) and the Kimi judge only $+1.21$, while Hard and Extreme success remain zero: with no adequate candidate branches to select among, selection cannot help on the hardest tasks, and an untrained policy cannot yet judge its own candidates. After SERA training, selection yields gains of $2.43$ (policy judge) and $2.22$ (Kimi judge) points that, crucially, extend to both harder levels. The trained policy judge now matches the external judge, indicating that SERA training both improves the candidate solutions available for selection and turns the policy into an effective self-evaluator. Together, these results support the SERA paradigm: training recursive execution and evaluation produces an agent that can further improve by generating and selecting alternative branches at inference.

\begin{table}[t]
\centering
\caption{Test-time scaling by bottom-up rubric-guided node selection on TextWorld-Sync, comparing Qwen3-4B-Instruct (no \method{} training) with the \method{}-trained policy. At each node we sample $N$ candidate resolutions and keep the one scored highest by the rubric, applied either by the policy itself (\emph{Policy judge}) or by an external frontier model (\emph{Kimi judge}); $N{=}1$ is a single rollout without selection. Success rates (\%) by difficulty with overall average; mean $\pm$ std over three runs. \textbf{Bold}/\underline{underline}: best/second-best per column.}
\label{tab:test_time_scaling}
\setlength{\tabcolsep}{5pt}
\scriptsize
\vspace{1pt}
\begin{tabular}{lcccc>{\columncolor{avgcol}}c}
\toprule
Selector & Easy & Medium & Hard & Extreme & Avg. \\
\midrule
\multicolumn{6}{@{}l}{\emph{Qwen3-4B-Instruct (no \method{} training)}} \\
Policy judge ($N{=}2$) & 54.67$\pm$1.67 & 11.62$\pm$4.06 & 0.00$\pm$0.00 & 0.00$\pm$0.00 & 16.57$\pm$1.42 \\
Kimi judge ($N{=}2$)   & 56.95$\pm$2.31 & 15.05$\pm$1.47 & 0.00$\pm$0.00 & 0.00$\pm$0.00 & 18.00$\pm$0.93 \\
No selection ($N{=}1$) & 54.19$\pm$0.59 & 12.95$\pm$0.33 & 0.00$\pm$0.00 & 0.00$\pm$0.00 & 16.79$\pm$0.21 \\
\midrule
\multicolumn{6}{@{}l}{\emph{\method{}-trained policy}} \\
Policy judge ($N{=}2$) & \underline{99.43}$\pm$0.49 & \underline{85.90}$\pm$1.15 & \textbf{55.71}$\pm$1.51 & \textbf{28.95}$\pm$0.33 & \textbf{67.50}$\pm$0.54 \\
Kimi judge ($N{=}2$)   & \textbf{99.90}$\pm$0.16 & \textbf{86.10}$\pm$0.92 & \underline{55.52}$\pm$2.29 & \underline{27.62}$\pm$0.72 & \underline{67.29}$\pm$0.69 \\
No selection ($N{=}1$) & 98.19$\pm$0.87 & 84.95$\pm$2.86 & 54.29$\pm$7.78 & 22.86$\pm$4.23 & 65.07$\pm$3.48 \\
\bottomrule
\end{tabular}
\vspace{-10pt}
\end{table}

\vspace{-10pt}
\section{Conclusion}
\vspace{-5pt}

We presented SERA, a method for training recursive language-model agents that casts training as three coupled capabilities: decomposing tasks into subtasks, solving them, and evaluating their outcomes. Each capability receives its own training signal, and the three are optimized by alternating updates on a single shared policy. 
The central finding is that training the policy to generate its own evaluation rubrics, through a ranking objective anchored to verified outcomes, is what drives the largest performance gains. Since evaluation and execution share weights, learning to judge subtask quality directly improves how the agent solves them. This trained rubric also reduces reliance on external supervision: the judge is needed only to train the rubric generator, and the resulting self-generated signal outperforms using the judge directly. Additionally, a leaf-coverage reward provides a direct signal for decomposition quality, with the largest gains on tasks requiring coordination among interdependent subtasks. Beyond training, the learned rubric serves as an inference-time selector for tree search, further improving task success without additional training.

Several limitations remain. We evaluate only Qwen3-4B on two synthetic benchmarks, so generalization to larger models and real-world agent tasks remains untested. Because the policy both writes and applies its own rubrics, self-evaluation could be gamed by scores that drift from genuine task success; we mitigate this by training rubric generation only against verified outcomes---a label-free dispersion reward instead degrades performance (Section~\ref{sec:ablation-study})---but robustness under sparser or noisier verification, and the residual dependence on verified outcomes to anchor rubric training (Table~\ref{tab:external_model_tokens}), remain open. Finally, the leaf-coverage reward treats all leaf nodes equally, and subtask-level importance weights may improve decomposition training.
Looking ahead, the rubric mechanism suggests learning rubrics that generalize across tasks, using the rubric to guide exploration during training, and extending self-evaluation to settings where even the root outcome is not exactly verifiable. Framing recursive-agent training as three coupled capabilities offers a useful foundation for this work.

\subsection*{AI use statement}

In this work, we used generative AI tools in a limited capacity for grammar correction and polishing of author-written text, and for coding assistance and debugging. All research ideas, method design, experimental design, result analysis, figures, and tables were produced entirely by the authors. Every AI-suggested text edit was reviewed and accepted or rejected by the authors, and all AI-assisted code was reviewed and tested by the authors before use. We take full responsibility for the final content of this work, including all text, claims, code, and artifacts.

\subsection*{Ethics statement}

This work trains language-model agents to decompose and solve tasks in simulated text environments. The benchmarks (TextCraft-Synth and TextWorld-Sync) are procedurally generated and do not involve human subjects, personal data, or real-world deployment. The self-evaluation mechanism enables agents to generate and score their own rubrics, which could in principle be misused to produce misleading self-assessments; we mitigate this by anchoring rubric training to externally verified outcomes and reporting all results against deterministic environment verification. We do not foresee significant negative societal impacts from this research.

\subsection*{Reproducibility statement}

All hyperparameters, training schedules, and evaluation protocols are reported in Section~\ref{sec:experiments} and Appendix~\ref{app:training-details}, including the alternating-update ratios, rollout group sizes, context lengths, and checkpoint steps for each method variant. The base model (Qwen3-4B-Instruct-2507) is publicly available. Both benchmarks use deterministic environment verification for all reported success rates. 
We plan to release the training code and the TextWorld-Sync benchmark upon publication.



\bibliography{iclr2027_conference}

@inproceedings{prasad2024adapt,
  title={Adapt: As-needed decomposition and planning with language models},
  author={Prasad, Archiki and Koller, Alexander and Hartmann, Mareike and Clark, Peter and Sabharwal, Ashish and Bansal, Mohit and Khot, Tushar},
  booktitle={Findings of the Association for Computational Linguistics: NAACL 2024},
  pages={4226--4252},
  year={2024}
}

@article{chi2025era,
  title={The Era of Agentic Organization: Learning to Organize with Language Models},
  author={Chi, Zewen and Dong, Li and Dong, Qingxiu and Hao, Yaru and Wu, Xun and Huang, Shaohan and Wei, Furu},
  journal={arXiv preprint arXiv:2510.26658},
  year={2025}
}

@inproceedings{schroeder2025thread,
  title={Thread: Thinking deeper with recursive spawning},
  author={Schroeder, Philip and Morgan, Nathaniel W and Luo, Hongyin and Glass, James},
  booktitle={Proceedings of the 2025 Conference of the Nations of the Americas Chapter of the Association for Computational Linguistics: Human Language Technologies (Volume 1: Long Papers)},
  pages={8418--8442},
  year={2025}
}

@article{zhang2025recursive,
  title={Recursive language models},
  author={Zhang, Alex L and Kraska, Tim and Khattab, Omar},
  journal={arXiv preprint arXiv:2512.24601},
  year={2025}
}

@article{gandhi2026recursive,
  title={Recursive Agent Optimization},
  author={Gandhi, Apurva and Chakraborty, Satyaki and Wang, Xiangjun and Kumar, Aviral and Neubig, Graham},
  journal={arXiv preprint arXiv:2605.06639},
  year={2026}
}

@article{tie2026autoresearch,
  title={Autoresearch ai: Towards ai-powered research automation for scientific discovery},
  author={Tie, Guiyao and Shi, Jiawen and Song, Dingjie and Huang, Yixiao and Sheng, Ziji and Zhou, Xueyang and Liu, Daizong and Zhou, Pan and Chen, Yongchao and Xu, Ran and others},
  journal={arXiv preprint arXiv:2605.23204},
  year={2026}
}

@misc{claudecode,
  title={Claude Code},
  author={{Anthropic}},
  year={2025},
  howpublished={\url{https://docs.anthropic.com/en/docs/claude-code}},
}

@misc{codex,
  title={Codex},
  author={{OpenAI}},
  year={2025},
  howpublished={\url{https://openai.com/index/introducing-codex/}},
}

@article{team2026kimi,
  title={Kimi k2. 5: Visual agentic intelligence},
  author={Team, Kimi and Bai, Tongtong and Bai, Yifan and Bao, Yiping and Cai, SH and Cao, Yuan and Chai, Ziwei and Charles, Y and Che, HS and Chen, Cheng and others},
  journal={arXiv preprint arXiv:2602.02276},
  year={2026}
}

@article{sun2025scaling,
  title={Scaling long-horizon llm agent via context-folding},
  author={Sun, Weiwei and Lu, Miao and Ling, Zhan and Liu, Kang and Yao, Xuesong and Yang, Yiming and Chen, Jiecao},
  journal={arXiv preprint arXiv:2510.11967},
  year={2025}
}

@article{dietterich2000hierarchical,
  title={Hierarchical reinforcement learning with the MAXQ value function decomposition},
  author={Dietterich, Thomas G},
  journal={Journal of artificial intelligence research},
  volume={13},
  pages={227--303},
  year={2000}
}

@article{nachum2018data,
  title={Data-efficient hierarchical reinforcement learning},
  author={Nachum, Ofir and Gu, Shixiang Shane and Lee, Honglak and Levine, Sergey},
  journal={Advances in neural information processing systems},
  volume={31},
  year={2018}
}

@inproceedings{lightman2024let,
  title={Let's verify step by step},
  author={Lightman, Hunter and Kosaraju, Vineet and Burda, Yuri and Edwards, Harrison and Baker, Bowen and Lee, Teddy and Leike, Jan and Schulman, John and Sutskever, Ilya and Cobbe, Karl},
  booktitle={International Conference on Learning Representations},
  volume={2024},
  pages={39578--39601},
  year={2024}
}

@article{zhang2026trca,
  title={TRCA: Transition-wise Rubric Credit Assignment for Long-horizon LLM Agents},
  author={Zhang, Huan and Chen, Mingju and Zhou, Dongxu and Lv, Can and Chang, Heng and Cui, Sen and Wu, Faguo and Zhou, Shiji},
  journal={arXiv preprint arXiv:2608.16156},
  year={2026}
}

@article{li2026and,
  title={What and When to Distill: Selective Hindsight Distillation for Multi-Turn Agents},
  author={Li, Xiaozhe and Lyu, Tianyi and Li, Yang and Ma, Yichuan and Li, Peiji and Li, Linyang and Guo, Qipeng and Lin, Dahua and Chen, Kai},
  journal={arXiv preprint arXiv:2605.19447},
  year={2026}
}

@article{tian2026arco,
  title={ARCO: Adaptive Rubric with Co-Evolution for Multi-Step LLM-Based Agents},
  author={Tian, Zihang and Zhang, Jingsen and Li, Rui and Bo, Xiaohe and Li, Yuanzi and Chen, Xu},
  journal={arXiv preprint arXiv:2606.21262},
  year={2026}
}

@inproceedings{liu2026openrubrics,
  title={Openrubrics: Towards scalable synthetic rubric generation for reward modeling and llm alignment},
  author={Liu, Tianci and Xu, Ran and Yu, Tony and Hong, Ilgee and Yang, Carl and Zhao, Tuo and Wang, Haoyu},
  booktitle={Proceedings of the 64th Annual Meeting of the Association for Computational Linguistics (Volume 1: Long Papers)},
  pages={17417--17437},
  year={2026}
}

@article{xu2026alternating,
  title={Alternating reinforcement learning for rubric-based reward modeling in non-verifiable llm post-training},
  author={Xu, Ran and Liu, Tianci and Dong, Zihan and Yu, Tony and Hong, Ilgee and Yang, Carl and Zhang, Linjun and Zhao, Tao and Wang, Haoyu},
  journal={arXiv preprint arXiv:2602.01511},
  year={2026}
}

@article{ding2026evorubrics,
  title={EvoRubrics: Dynamic Rubrics as Rewards via Adversarial Co-Evolution for LLM Reinforcement Learning},
  author={Ding, Hongxin and Huang, Baixiang and Fang, Yue and Liao, Weibin and Li, Zheng and Zhang, Jinyang and Wu, Zhijing and Zhao, Junfeng and Wang, Yasha},
  journal={arXiv preprint arXiv:2606.23038},
  year={2026}
}

@inproceedings{gao2025scaling,
  title={Scaling and evaluating sparse autoencoders},
  author={Gao, Leo and Dupre la Tour, Tom and Tillman, Henk and Goh, Gabriel and Troll, Rajan and Radford, Alec and Sutskever, Ilya and Leike, Jan and Wu, Jeffrey},
  booktitle={International Conference on Learning Representations},
  volume={2025},
  pages={26721--26754},
  year={2025}
}

@article{wang2026reproducing,
  title={Reproducing, Analyzing, and Detecting Reward Hacking in Rubric-Based Reinforcement Learning},
  author={Wang, Xuekang and Hao, Zhuoyuan and Hou, Shuo and Peng, Hao and Li, Juanzi and Wang, Xiaozhi},
  journal={arXiv preprint arXiv:2606.04923},
  year={2026}
}

@article{yang2026rubric,
  title={Rubric Dropout: A Simple Way to Mitigate Reward Hacking in Rubric-as-Reward RL},
  author={Yang, Minglai and Guo, Xinyu and Tyagi, Utkarsh and Zhang, Mian and Dumitru, Razvan and Hou, Sunjie and He, Yunzhong and Zhang, Daniel Yue and Liu, Ying},
  journal={arXiv preprint arXiv:2608.11669},
  year={2026}
}

@article{minimax2025m1,
  title={Minimax-m1: Scaling test-time compute efficiently with lightning attention},
  author={Chen, Aili and Li, Aonian and Gong, Bangwei and Jiang, Binyang and Fei, Bo and Yang, Bo and Shan, Boji and Yu, Changqing and Wang, Chao and Zhu, Cheng and others},
  journal={arXiv preprint arXiv:2506.13585},
  year={2025}
}

@article{shao2024deepseekmath,
  title={Deepseekmath: Pushing the limits of mathematical reasoning in open language models},
  author={Shao, Zhihong and Wang, Peiyi and Zhu, Qihao and Xu, Runxin and Song, Junxiao and Bi, Xiao and Zhang, Haowei and Zhang, Mingchuan and Li, YK and Wu, Yang and others},
  journal={arXiv preprint arXiv:2402.03300},
  year={2024}
}

@article{yang2025qwen3,
  title={Qwen3 technical report},
  author={Yang, An and Li, Anfeng and Yang, Baosong and Zhang, Beichen and Hui, Binyuan and Zheng, Bo and Yu, Bowen and Gao, Chang and Huang, Chengen and Lv, Chenxu and others},
  journal={arXiv preprint arXiv:2505.09388},
  year={2025}
}

@inproceedings{jansen2023textworldexpress,
  title={Textworldexpress: Simulating text games at one million steps per second},
  author={Jansen, Peter and C{\^o}t{\'e}, Marc-Alexandre},
  booktitle={Proceedings of the 17th Conference of the European Chapter of the Association for Computational Linguistics: System Demonstrations},
  pages={169--177},
  year={2023}
}

@article{team2026kimi-k3,
  title={Kimi k3: Open frontier intelligence},
  author={Team, Kimi and Bai, Tongtong and Bai, Yifan and Bao, Yiping and Cai, Jianfeng and Cai, Xinyuan and Cao, Peizhou and Cao, Yuxuan and Chai, Ziwei and Charles, Y and others},
  journal={arXiv preprint arXiv:2607.24653},
  year={2026}
}

@misc{RubricEM,
      title={RubricEM: Meta-RL with Rubric-guided Policy Decomposition beyond Verifiable Rewards}, 
      author={Gaotang Li and Bhavana Dalvi Mishra and Zifeng Wang and Jun Yan and Yanfei Chen and Chun-Liang Li and Long T. Le and Rujun Han and George Lee and Hanghang Tong and Chen-Yu Lee and Tomas Pfister},
      year={2026},
      eprint={2605.10899},
      archivePrefix={arXiv},
      primaryClass={cs.CL},
      url={https://arxiv.org/abs/2605.10899}, 
}
\bibliographystyle{iclr2027_conference}

\appendix
\newpage

\section{Recursive Execution and Supervision}
\label{app:rao-rubric}

\begin{figure}[H]
    \centering
    \includegraphics[width=\textwidth]{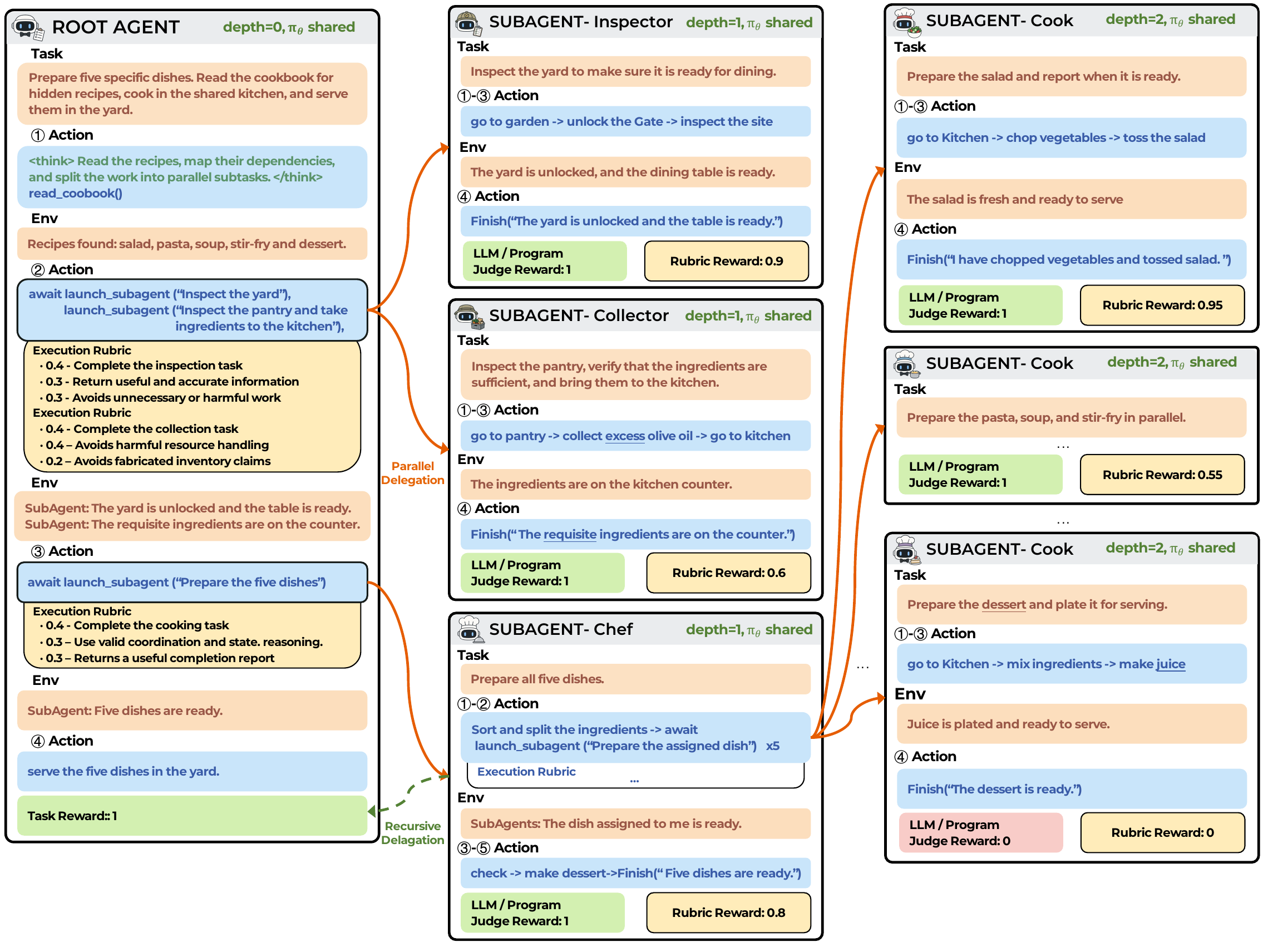}
    \caption{Recursive execution and trajectory scoring with a shared policy.
    Parents delegate subtasks to child agents and commit rubrics before child execution.
    The scoring sources shown have different roles across methods: external outcomes directly reward execution in RAO, whereas policy-generated rubric scores reward subtask execution in SERA.
    SERA uses external subtask labels to train rubric generation through a separate ranking objective.}
    \label{fig:rao-rubric-system}
\end{figure}

Figure~\ref{fig:rao-rubric-system} illustrates the recursive execution structure shared by SERA and its baselines.
The root agent delegates subtasks to child instances of the same policy, which can delegate further or solve their assigned tasks directly.
Independent branches can run in parallel, while parents incorporate returned results into their own execution.
In SERA, each parent generates a subtask-specific rubric before the child runs and freezes it throughout that child's execution.
After completion, the policy applies this rubric to the child's trajectory and final environment state, producing its execution reward.

The rubric scorer and external verifier serve different purposes in this framework.
RAO uses externally supplied subtask outcomes directly as execution rewards; SERA uses them to train rubric generation on counterfactual rollouts.
These outcomes come from a program verifier on TextCraft-Synth and an external LLM judge on TextWorld-Sync.
SERA's ordinary subtask execution instead receives the policy's own rubric scores, while the root retains its environment-verified outcome.
The leaf-coverage reward separately credits delegation using the completed tree and root outcome, without additional subtask labels.
Thus, the three signals train execution, decomposition, and evaluation through distinct updates, as specified in Section~\ref{sec:method}.

\section{Training and Evaluation Settings}
\label{app:training-details}

\subsection{Shared Optimization and Evaluation Protocol}

We use Qwen3-4B-Instruct-2507~\citep{yang2025qwen3} as the base policy and fine-tune all parameters with Adam in the RL experiments.
The recursive RL methods share the CISPO objective~\citep{minimax2025m1}, leave-one-out advantage estimation, and inverse-frequency depth weighting from Section~\ref{sec:method-objective}.
Each optimizer update uses eight root tasks, with $K=8$ independently sampled trees per root task.
The root agent and each child have a budget of 20 environment steps, with a maximum recursion depth of three.
Table~\ref{tab:shared_training_settings} reports the detailed optimization settings, rollout budgets, and evaluation configuration.

\begin{table}[H]
\scriptsize
\centering
\caption{SERA training and evaluation settings. Values shown as ranges are due to small entrance-specific prompt/completion budget differences within the same benchmark.}
\label{tab:shared_training_settings}
\resizebox{\textwidth}{!}{%
\begin{tabular}{lcc}
\toprule
\textbf{Setting} & \textbf{TextCraft-Synth} & \textbf{TextWorld-Sync} \\
\midrule

Hardware
& \multicolumn{2}{c}{2 $\times$ 8 H200 GPUs} \\

Batch size / group size
& \multicolumn{2}{c}{8 / 8 root tasks} \\

Root / SubAgent steps
& \multicolumn{2}{c}{20 / 20} \\

Maximum recursion depth
& \multicolumn{2}{c}{3} \\

Training temperature
& \multicolumn{2}{c}{1.0} \\

Optimizer
& \multicolumn{2}{c}{Adam, $\mathrm{lr}=3\times10^{-6}$} \\

Weight decay / gradient clipping
& \multicolumn{2}{c}{0.017 / 1.0} \\

Loss / advantage estimator
& \multicolumn{2}{c}{CISPO / LOO + depth weighting} \\

Rubric-generation rollout group size $B$
& \multicolumn{2}{c}{8} \\

\midrule

Training split
& 852 Medium tasks
& 900 Medium / 150 Hard / 450 Extreme \\

Context length
& 10,240 tokens
& 13,312 tokens \\

Prompt / completion budget
& 9,728 / 512 tokens
& 10,240 / 3,072 tokens \\

Evaluation set
& 632 tasks: 147 / 213 / 136 / 136
& 1,400 tasks: 350 per difficulty \\

\bottomrule
\end{tabular}%
}
\end{table}

\subsection{Reward Sources and Alternating Updates}

Table~\ref{tab:training_method_schedule} summarizes the reward configurations and update ratios using the method names from Table~\ref{tab:main_compare}.
Here, D denotes decomposition training with the leaf-coverage reward, and RT denotes rubric-generation training with the verified-outcome ranking reward.
Execution updates supervise solving and delegation tokens; D updates supervise delegation tokens alone, and RT updates supervise rubric-generation tokens alone.
Each optimizer step activates one objective on the shared policy, rather than summing the three raw rewards.

For rubric-based execution, the policy both generates subtask-specific criteria and scores completed trajectories against them.
Generating or applying a rubric does not itself constitute RT; RT explicitly optimizes the rubric-generation action using verified counterfactual outcomes.
Accordingly, SERA w/o D\,\&\,RT uses rubric scores without explicitly training the rubric generator.
On TextWorld-Sync, the external LLM judge supplies binary subtask labels, whereas the policy rubric scorer supplies continuous execution scores.
These training signals are distinct from the deterministic root verifier used to report benchmark success.

We allocate most updates to execution, which directly trains task completion throughout the recursive tree.
As the policy changes, periodic D and RT updates adjust its delegation decisions and the criteria that supply its execution rewards.
Their data requirements differ: D uses ordinary execution trees, whereas RT requires additional counterfactual rollouts and external labels.
We therefore keep RT updates infrequent while interleaving them with execution training, allowing the evaluator to adapt to the evolving rollout distribution.

SERA uses an execution:D:RT ratio of $16:2:2$, reserving four of every twenty updates for decomposition and rubric generation.
SERA w/o D uses $16:0:2$, retaining periodic rubric training while assigning the remaining updates to execution.
RAO + D uses $12:4:0$, devoting a larger fraction to delegation updates, which require neither counterfactual sampling nor rubric scoring.
The execution-only variants use $1:0:0$ and optimize execution at every step.
These ratios control how often each objective updates the shared policy; they are not coefficients applied to a combined reward.

All recursive methods use the same total optimizer-update budget within each benchmark, with D and RT occupying part of that budget.
The comparisons thus hold total update counts fixed but use the stage allocations in Table~\ref{tab:training_method_schedule}.

\begin{table}[H]
\footnotesize
\centering
\caption{Subtask execution rewards, alternating update ratios, and uses of external supervision. D denotes decomposition training; RT denotes rubric-generation training. Ratios follow execution:D:RT, with one objective active per update. External subtask labels come from the program verifier on TextCraft-Synth and the LLM judge on TextWorld-Sync.}
\label{tab:training_method_schedule}
\setlength{\tabcolsep}{5pt}
\resizebox{\textwidth}{!}{%
\begin{tabular}{llll}
\toprule
Method & Subtask execution reward & Update ratio & External subtask labels used for \\
\midrule
RAO & Binary outcome & $1:0:0$ & Execution rewards \\
RAO + D & Binary outcome & $12:4:0$ & Execution rewards \\
RAO + RT & Binary outcome & $16:0:2$ & Execution rewards and Rubric training \\
\midrule
\method{} w/o D\,\&\,RT & Policy rubric score & $1:0:0$ & None \\
\method{} w/o D & Policy rubric score & $16:0:2$ & Rubric training \\
\method{} & Policy rubric score & $16:2:2$ & Rubric training \\
\bottomrule
\end{tabular}%
}
\end{table}

\subsection{Training Runtime}
\label{app:runtime}

Table~\ref{tab:runtime} reports wall-clock training time to the evaluated checkpoint for each recursive method.
All runs use identical hardware (2\,$\times$\,8 H200 GPUs) and the shared settings of Table~\ref{tab:shared_training_settings}, so we read the absolute times only as relative comparisons within each benchmark rather than as hardware-independent costs.
Because all recursive methods share the same total optimizer-update budget, SERA's decomposition and rubric-training stages displace execution updates rather than adding to them; runtime differences therefore reflect the per-update cost of each stage together with the rollout cost of each benchmark.

On TextCraft-Synth, SERA is the fastest multi-agent method at $24$h$06$m, below RAO's $29$h$36$m and well below the rubric-at-every-step ablation \method{} w/o D\,\&\,RT at $47$h$02$m, while also being the most accurate (Table~\ref{tab:main_compare}).
Concentrating rubric computation into periodic rubric-generation training, rather than scoring every execution step against a rubric, is therefore what keeps SERA's cost low on this benchmark.

On TextWorld-Sync, SERA instead takes $24$h$14$m versus RAO's $13$h$08$m.
The two benchmarks diverge because their rollouts differ in cost: independent crafting branches in TextCraft admit parallel execution, whereas TextWorld's shared-state dependencies force sequential interaction.
The counterfactual rollouts used for rubric-generation training must also traverse these dependencies, and local rubric generation and scoring add per-step overhead even as external judge calls fall (Table~\ref{tab:external_model_tokens}).
SERA thus trades additional training time on TextWorld for a $13.14$-point average accuracy gain over RAO.

\begin{table}[H]
\centering
\caption{Training wall-clock time to the evaluated checkpoint on 2\,$\times$\,8 H200 GPUs, under the shared settings of Table~\ref{tab:shared_training_settings}. Times are implementation- and hardware-dependent and are meaningful only as relative comparisons within a benchmark. Prompting-only baselines involve no training and are omitted. D\,=\,Decomposition reward, RT\,=\,Rubric Training.}
\label{tab:runtime}
\small
\begin{tabular}{lcc}
\toprule
Method & TextCraft & TextWorld \\
\midrule
\multicolumn{3}{@{}l}{\emph{Single-Agent}} \\
CISPO & 29h\,06m & 7h\,43m \\
GRPO  & 28h\,17m & 8h\,06m \\
\midrule
\multicolumn{3}{@{}l}{\emph{Multi-Agent}} \\
RAO                    & 29h\,36m & 13h\,08m \\
RAO + D                & 24h\,57m & 11h\,06m \\
RAO + RT               & 24h\,56m & 27h\,31m \\
\method{} w/o D\,\&\,RT & 47h\,02m & 16h\,14m \\
\method{} w/o D         & 34h\,43m & 23h\,48m \\
\method{}               & 24h\,06m & 24h\,14m \\
\bottomrule
\end{tabular}
\end{table}

\section{TextWorld-Sync Dataset Construction}
\label{apd:text_world_dataset}

TextWorld-Sync extends the CookingWorld environment in TextWorldExpress~\citep{jansen2023textworldexpress} to test recursive delegation under shared-state constraints.
Its design separates local subtask completion from joint task success, complementing the independently verifiable crafting subtasks in TextCraft-Synth.

\subsection{Shared World and Agent-Local Contexts}

The standard TextWorldExpress interface exposes one mutable game through a single avatar.
We retain the CookingWorld dynamics and introduce a coordinator that lets recursive agents interact with one shared world.
Objects, containers, inventory, tools, doors, recipe progress, and task score are shared across agents.
Each agent maintains its own logical location, dialogue history, observation stream, and step budget.
Model calls may run concurrently, but environment reads and state transitions are serialized under a lock and committed atomically.
Consequently, an item acquired by one agent becomes unavailable to others, while their locations and histories remain separate.
This distinction permits concurrent planning while preserving consistent state updates and exposing dependencies between agents' actions.

\subsection{Procedural Tasks and Coordination Structures}

An original CookingWorld instance specifies one recipe, its required ingredients, and a target meal.
TextWorldExpress procedurally samples recipes, layouts, object placements, and difficulty parameters.
We extend this generator to sample multiple disjoint recipes per seed, assign ingredients to dishes, and introduce one of seven coordination structures.
The complete instance is recorded in a fixed manifest and reconstructed at runtime by a multi-dish adapter.
Independent dish branches can be delegated in parallel, subject to the instance's resource and dependency constraints.

Table~\ref{tab:textworld_sync_families} summarizes the seven task families and the coordination decisions they require.
They range from independent branches to shared-resource conflicts and nested dependencies, providing different conditions for evaluating decomposition quality.
The deterministic root evaluator reads structured environment state and requires every ingredient to be collected and correctly processed, and every dish prepared and consumed.
Completing an individual dish yields partial progress but does not terminate the root task.
This root-level check does not provide labels for arbitrary policy-generated subtasks; SERA therefore uses an external judge for rubric-training labels on this benchmark.

\begin{table}[H]
\small
\centering
\caption{Coordination structures in TextWorld-Sync. The seven task families vary branch dependencies, workload allocation, resource constraints, and recursion. Every family uses a shared world and requires completion of all dishes for root-task success.}
\label{tab:textworld_sync_families}
\begin{tabular}{@{}>{\raggedright\arraybackslash}p{0.24\textwidth}>{\raggedright\arraybackslash}p{0.71\textwidth}@{}}
\toprule
Task family & Structure and coordination requirement \\
\midrule
Fork--join & Independent dish branches converge on shared preparation and consumption, requiring their results to be coordinated. \\
Critical path & One branch has a longer workload; launching it early reduces delays before the branches can join. \\
Load balanced & Branch workloads differ, requiring work to be distributed across agents. \\
Resource constrained & Inventory capacity, available tools, or a shared step budget limit concurrent work. \\
Dependency gate & A branch can proceed only after a prerequisite, such as opening a door, has been satisfied. \\
Conflict & Branches compete for a unique shared tool, such as a knife or stove, requiring coordinated access. \\
Hierarchical & Dish-level tasks contain ingredient-level subtasks, supporting delegation across multiple recursion depths. \\
\bottomrule
\end{tabular}
\end{table}

\subsection{Dataset Splits and Difficulty Levels}

Tasks are assigned to Easy, Medium, Hard, or Extreme according to the number and scale of their constraints.
The fixed TextWorld-Sync split contains 1,500 training tasks, 200 validation tasks, and 1,400 test tasks.
Training comprises 900 Medium, 150 Hard, and 450 Extreme tasks; validation and test contain 50 and 350 tasks per difficulty, respectively.
For comparison, TextCraft-Synth training uses the fixed subset of 852 Medium tasks, with evaluation on 632 tasks spanning all four levels.
Thus, TextCraft-Synth tests transfer from Medium training tasks across recipe depths, whereas TextWorld-Sync training covers several levels of coordination difficulty.
Table~\ref{tab:shared_training_settings} reports the corresponding evaluation-set sizes by difficulty.

\end{document}